\documentclass[11pt,a4paper]{article}
\usepackage[hyperref]{emnlp2018}
\usepackage{times}
\usepackage{latexsym}
\usepackage{multirow}
\usepackage{graphicx}
\usepackage{multirow}
\usepackage{amsmath}
\usepackage{url}
\usepackage{color}
\usepackage{amssymb}
\usepackage{algorithmic}
\usepackage{algorithm}
\usepackage{url}

\aclfinalcopy % Uncomment this line for the final submission
\def\fndaff{$^\dagger$}
\def\sstaff{$^\mathsection$}

\title{Multi-Level Structured Self-Attentions for Distantly \\ Supervised Relation Extraction}

\author{{Jinhua Du\fndaff, Jingguang Han\sstaff, Andy Way\fndaff, Dadong Wan\sstaff} \\
  \fndaff ADAPT Centre, School of Computing, Dublin City University, Ireland \\
  \sstaff Accenture Labs Dublin, Ireland \\
     {\{jinhua.du, andy.way\}@adaptcentre.ie} \\ 
     {\{jingguang.han, dadong.wan\}@accenture.com}}

\date{}

\begin{document}
\maketitle
\begin{abstract}
  Attention mechanisms are often used in deep neural networks for distantly supervised relation extraction (DS-RE) to distinguish valid from noisy instances. However, traditional 1-$D$ vector attention models are insufficient for the learning of different contexts in the selection of valid instances to predict the relationship for an entity pair. % Distant supervision is an efficient method to automatically annotate large-scale data using knowledge bases for relation extraction (RE). However, it suffers from a severe mis-labelling problem. To alleviate this issue, multi-instance learning (MIL) is often used to distinguish noisy from informative instances. 
To alleviate this issue, we propose a novel multi-level structured (2-$D$ matrix) self-attention mechanism for DS-RE in a multi-instance learning (MIL) framework using bidirectional recurrent neural networks. In the proposed method, a structured word-level self-attention mechanism learns a 2-$D$  matrix where each row vector represents a weight distribution for different aspects of an instance regarding two entities. Targeting the MIL issue, the structured sentence-level attention %is expanded from a 1-$D$ vector to a structured 2-$D$ matrix, which 
  learns a 2-$D$ matrix where each row vector represents a weight distribution on selection of different valid instances.   %select the most likely sentence for each entity pair
Experiments conducted on two publicly available DS-RE datasets show that the proposed framework with a multi-level structured self-attention mechanism significantly outperform state-of-the-art baselines in terms of PR curves, P@N and F1 measures.% in terms of Precision-Recall curves, P@N and Macro F1 measures.
\end{abstract}

\section{Introduction}
Relation extraction is a fundamental task
in information extraction (IE), which 
studies the issue of predicting semantic relations between pairs of entities in a sentence~\cite{Zelenko2003,Bunescu2005,guodong2005}. %It can be formalised as: given a sentence $S$ with the annotated pairs of nominals $e_1$ and $e_2$, the goal is to identify the relation $r$ from a predefined relation set $\mathbb{R}$ for the entity pair $(e_1, e_2)$, written in the form of a triple $(e_1, r, e_2)$ or $(e_2, r, e_1)$~\cite{hendrickx2009semeval}. %For example, \emph{fire} and \emph{fuel} are identified as a \emph{Cause-Effect} relationship in the sentence ``The $[$fire$]_{e_1}$  inside WTC was caused by exploding $[$fuel$]_{e_2}$'', which can be denoted as $(\emph{fuel}, \emph{Cause-Effect}, \emph{fire})$. %Relation extraction
%is the key component for building semantic knowledge
%graphs~\cite{bordes:2014,nickel:2016,augenstein:2016}, and it is of crucial significance to some natural language processing (NLP) applications such as
%structured search \cite{yahya:2016}, sentiment analysis, question answering \cite{fader:2014,yih-he-meek:2014:P14-2},
%and text summarisation.
One crucial problem in RE is the relative lack of large-scale, high-quality labeled data. In recent years, one commonly used and effective technique for dealing with this challenge is %{\color{red} a -} 
the distant supervision method via knowledge bases (KBs)~\cite{Mintz:2009,Riedel:2010,Hoffmann:2011}, which assumes that if one entity pair appearing in some sentences can be observed in a KB with a certain relationship, then these sentences will be labeled as the context of this entity pair and this relationship. The distant %{\color{red} distance} 
supervision strategy is an effective and efficient method for automatically labeling large-scale training data. However, it also introduces a severe  mislabelling problem due to the fact that a sentence that mentions two entities does not necessarily express their relation in a KB~\cite{Surdeanu:2012,Zeng2015}. 

Plenty of research work has been proposed to deal with distantly supervised data and has achieved significant progress, especially with the rapid development of deep neural networks (DNN) for relation extraction in recent years~\cite{Zeng2014,Zeng2015,Lin2016acl,Lin2017acl,WangLinlin2016,Zhoupeng2015,GuoliangJi,Linyi17,Zeng2017}. 
%Deep neural network-based (
DNN models under an MIL framework for DS-RE have become state-of-the-art, replacing statistical methods, such as  feature-based and graphical models ~\cite{Riedel:2010,Hoffmann:2011,Surdeanu:2012}. In the MIL framework for %{\color{red} distance} 
distantly supervised RE, each entity pair %with one or more relationships, defined as the single-label problem or multi-label problem, 
often has multiple instances where some are noisy and some are valid. %We define all the instances mentioning both $e_1$ and $e_2$ in a tuple $r(e_1, e_2)$ to constitute a bag $\mathcal{G}$ with a relation $r$. An effective classifier should have the capability to select informative sentences for a correct relation prediction. %, and the relation $r$ is the label of the bag. In the bag $\mathcal{G}$, some instances are noise and some are informative. %The labels of the bags are known, but the labels of the instances in the bags are unknown. In the learning process, the uncertainty of instance labels can be taken into account, such as selecting one instance with highest probability as the valid instance, this alleviates the wrong label problem~\cite{Zeng2015,GuoliangJi}. 
The attention mechanism in DNNs, such as convolutional (CNN) and recurrent neural networks (RNN), 
is an effective way to select valid instances by learning a weight distribution over multiple instances. However, there are two important representation learning problems in DNN-based distantly supervised RE: (1) \textbf{Problem I}: entity pair-targeted context representation learning from an instance; and (2) \textbf{Problem II}: valid instance selection representation learning over multiple instances. The former can use a word-level attention mechanism to learn a weight distribution on words %at different time steps 
and then a weighted sentence representation regarding two entities; the latter can employ a sentence-level attention mechanism to learn a weight distribution on multiple instances so that valid sentences with higher weights
can be focused and selected, and noisy instances with lower weights 
are suppressed. 

Both the word-level and sentence-level attention mechanisms in previous work on the RE task are simple 1-$D$ vectors which are learned using the hidden states of the RNN, or via pooling from either the RNNs' hidden states or convolved $n$-grams~\cite{Zeng2014,Zeng2015,Zhoupeng2015,WangLinlin2016,GuoliangJi,Linyi17}. 
The deficiency of the 1-$D$ attention vector is that %the highest weighted word or instance contributes more to the target, i.e. 
it only focuses on one or a small number of aspects of the sentence, or one or a small number of instances~\cite{linzhouhan2017}, with the result that different semantic aspects of the sentence, or different multiple valid sentences are ignored, and cannot be utilised.  %However, in the distant supervised RE task, the entity pair generally depends on different aspects of context and relies on many different valid instances for a certain relation. 

Inspired by the structured self-attentive sentence embedding in~\citet{linzhouhan2017}, we propose a novel multi-level structured (2-$D$) self-attention mechanism (MLSSA) in a bidirectional LSTM-based (BiLSTM)~\cite{Hochreiter:1997} MIL framework to alleviate two \textbf{problems} in the distantly supervised RE. Regarding \textbf{Problem I}, we propose a 2-$D$ matrix-based word-level attention mechanism, which contains multiple vectors, each focusing on different aspects of the sentence for better context representation learning. % that the system should focus on. 
In terms of \textbf{Problem II}, we propose a 2-$D$ sentence-level attention mechanism for multiple instance learning, where it contains multiple vectors, each focusing on different valid instances for a better sentence selection. ``{\bf structured}'' indicates that the weight vectors in the learned 2-$D$ matrix try to construct a structural dependency relationship by learning different weight distributions for different contexts or instances given the entity pair. We can see that our structured attention mechanism is different from that in~\citet{kim2017} which incorporates richer structural distributions and are simple extensions of the basic attention procedure.
We verify the proposed framework on two distantly supervised RE datasets, namely the New York Times (NYT) dataset~\cite{Riedel:2010} and the DBpedia Portuguese dataset~\cite{Batista:2013}. Experimental results show that our MLSSA framework significantly outperforms state-of-the-art baseline systems in terms of different evaluation metrics. 

The main contributions of this paper include: (1) we propose a novel multi-level structured (2-$D$) self-attention mechanism for DS-RE which can make full use of input sequences to learn different contexts, without integrating extra resources;
(2) we %extend the traditional 1-$D$ word-level attention to 
propose a 2-$D$ matrix-based word-level attention %learn different aspects of the sentence 
for better context representation learning targeting two entities; 
(3) we propose a 2-$D$ sentence-level attention mechanism over multiple instances to select different valid instances; and (4) we verify the proposed framework on two publicly available distantly supervised datasets.

%The rest of the paper is organised as follows. In Section~\ref{relatedwork}, related work dealing with the distant supervised data is introduced. Section~\ref{ourmodel} details the proposed multi-level and multi-scale attention model for the RE task. In Section~\ref{experiments}, experimental results of the proposed method on one widely-used data set are reported, and analysis of results is carried out. Section~\ref{conclusion} gives our conclusion and points out future work. 

\section{Related Work}\label{relatedwork}
Most existing work on distant supervision data mainly focuses on denoising the data under the MIL strategy by learning a valid sentence representation or features, and then selecting one or more valid instances for relation classification~\cite{Riedel:2010,Hoffmann:2011,Surdeanu:2012,Zeng2015,Lin2016acl,Lin2017acl,Zhoupeng2015,GuoliangJi,Zeng2017, Linyi17}. 

\citet{Riedel:2010} and \citet{Surdeanu:2012} use a graphical model and MIL to select the valid sentences and classify the relations. However, these models are based on statistical methods and feature engineering, i.e. extracting sentence features using other NLP tools. \citet{Zeng2015} proposed a piece-wise CNN (PCNN) method to automatically learn sentence-level features and select one valid instance for the relation classification. The one-sentence-selection strategy does not make full use of the supervision information among multiple instances. 

\citet{Lin2016acl} and \citet{GuoliangJi} introduce an attention mechanism to the PCNN-based MIL framework to select informative sentences, which outperforms all baseline systems on the NYT data set. However, their attention mechanism is only a sentence-level model without incorporating word-level attention. \citet{Zhoupeng2015} introduce a word-level attention model to the BiLSTM-based MIL framework and obtain significant improvements on the SemEval2010~\cite{hendrickx2009semeval} data set.   \citet{WangLinlin2016} extend the single word-level attention model to multiple word levels in CNNs to discern patterns in heterogeneous contexts of the input sentence, and achieve best performance on the SemEval2010 data set. %However, these their multi-level mechanism includes only different word-level attentions for 
However, these two works were not targeting the %did not validate their systems on the 
distantly supervised RE problem. %dataset. Furthermore, word-level attention itself does not have the capability of selecting valid instances regarding the distantly supervised RE.%because it capture both entity-specific attention at the input level and relation-specific pooling attention (secondary attention with respect to the target relations). This allows it
% detecting more subtle cues of the heterogeneous structure in the input sentences, enabling it to automatically learn which parts are relevant for a given classification. They did not verify their method on the NYT distance supervised data either. 

\citet{Linyi17} experiment with word-level and sentence-level attention models in the bidirectional RNN on the NYT dataset on the basis of the open source DS-RE system,\footnote{\url{https://github.com/frankxu2004/TensorFlow-NRE}} and verify that a two-level attention mechanism achieves best performance compared to PCNN/CNN models. Both the word-level and sentence-level attention models are 1-$D$ vectors.
 
From previous work, we can see that the attention mechanism in DNNs has made significant progress on the RE task. However, both word-level and sentence-level attention models are 
still based on 1-$D$ vectors which have the following insufficiencies: %, which is a probability distribution over different time-step words or multiple instances. 
%The insufficiencies of 1-$D$ vector-based attention include: 
(1) although the 1-$D$ attention model can learn weights for different contexts, it only focuses on one or very few aspects of a single sentence~\cite{linzhouhan2017}, or one or very few instances; %, so the highest weighted word or instance contributes more to the target. 
(2) in order to allow the attention mechanism to learn more aspects of the sentence, or different instances, extra knowledge needs to be integrated, such as the work in~\citet{GuoliangJi} and \citet{Lin2017acl}. The former integrates entity descriptions generated from Freebase and Wikipedia as supplementary background knowledge to disambiguate the entity. The latter introduces a multilingual framework which employs a monolingual attention mechanism to utilize the information within monolingual texts, and further uses a cross-lingual attention mechanism to consider the information consistency and complementarity among cross-lingual texts. However, extra resources are difficult to obtain in many practical scenarios.

In order to alleviate the burden of integrating extra knowledge, and make full use of the input sentence (i.e. learning different aspects of context and focusing on different valid instances), we propose a multi-level structured self-attention mechanism in a BiLSTM-based MIL framework without integrating extra resources. %In our framework, we design a word-level attention for the sentence embedding learning, and the sentence-level attention for the instance selection. Furthermore, the 1-$D$ vector attention is extended to a 2-$D$ matrix to learn more aspects of the contexts for a better semantic representation. The experiments conducted on the NYT data set show the effectiveness of the proposed model.%The scale of word-level attention is a 2-$D$ structure of matrix or multiple vector representations, where each vector represents different aspects of the sentence that the system should focus on. Following the multiple sentence representations from the word-level attention, we propose different scales of the sentence-level attention mechanism, i.e. a 1-$D$ vector or 2-$D$ matrix. Theoretically, the 2-$D$ sentence-level attention can focus on different valid instances compared to the 1-$D$ attention. Under different settings of the proposed method, experimental results on the distant supervised data set show that the combination of 2-$D$ word-level attention and 1-$D$ sentence-level attention is more effective and robust, and significantly outperforms all the baseline systems. 

\section{Approach}\label{ourmodel}
\begin{figure*}[t]
\centering
\includegraphics[scale=0.34]{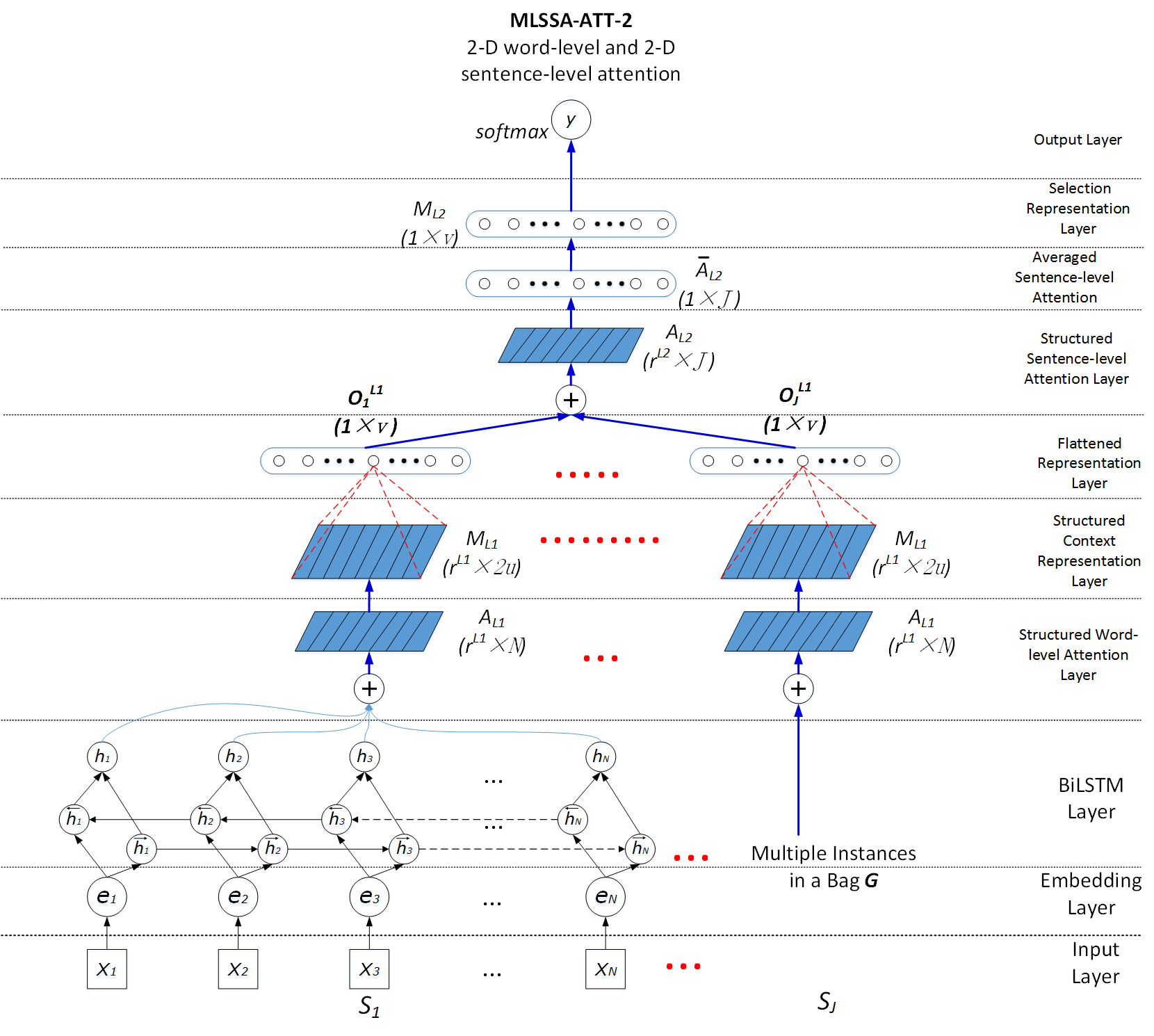}
\caption{Multi-level structured self-attention framework for distantly supervised RE}
\label{model_arch}
\end{figure*}
The distantly supervised RE can be formalised as follows: given an entity pair $(e_1, e_2)$, a bag $\mathcal{G}$ containing $\mathcal{J}$ instances, and the relation label $r$ for $\mathcal{G}$, the goal of the training process is to denoise these instances by selecting valid candidates based on $r$, and the goal of the testing process is to denoise multiple instances by selecting valid candidates to predict the relation $r$ for $\mathcal{G}$.

To alleviate the aforementioned two problems, improving the following two representation learning issues is clearly important for a DNN-based RE classifier: 
\begin{itemize}
\item {\it Entity pair-targeted context representation}: The model should have the capability to learn a better context representation from the input sentence targeting the entity pair; 
\item {\it Instance selection representation}: The model should have the capability to learn a better weight distribution over multiple instances to select valid instances regarding an entity pair.
\end{itemize}

Motivated by these two issues, we propose a multi-level structured self-attention framework.

\subsection{Architecture}
The proposed %multi-level and multi-scale self-attention 
framework consists of three parts as shown in Figure~\ref{model_arch}. The first part includes the input layer, embedding layer and BiLSTM layer which transform the input sequence at different time steps to LSTM hidden states. 

The second part implements the entity pair-targeted context representation learning, including: 
\begin{itemize}
\item {\it a structured word-level self-attention layer}: this generates a set of summation weight vectors (or a 2-$D$ matrix) taking the LSTM hidden states as input. Each vector in the 2-$D$ matrix represents the weights for different aspects of the input sentence. % for an input sentence. 
\item {\it a structured context representation layer}:  the weight vectors learned by the 2-$D$ word-level self-attention are dotted with the BiLSTM hidden states. Accordingly, a 2-$D$ matrix or a set of weighted LSTM hidden state vectors, denoted as ``$M_{L1}$'' in Figure~\ref{model_arch}, is obtained. Each weighted vector represents a sentence embedding reflecting a different aspect of the sentence targeting the entity pair. By this means, a dependency parsing-like structure of the input sentence can be constructed, obtaining different semantic representations of the sentence for the two entities in question. 
\item {\it a flattened representation layer}: this concatenates each vector in the 2-$D$ matrix of the sentence embedding to one vector. Then, the flattened vector connects to a 1-layer multi-layer perceptron (MLP) with ReLU activation function, generating an aggregated sentence representation.  %by considering all different aspects of the contexts from the structured representation layer.

\end{itemize}

The first and second parts operate on the single instance level, i.e. given a bag $\mathcal{G}$ and feeding each instance into the framework, the structured word-level self-attention mechanism will construct $J$ individual structured sentence representations corresponding to $J$ input instances.

The third part targets the instance selection representation learning issue, and operates on the bag level, i.e. considering weighted context representations of all instances in the bag $\mathcal{G}$ and learning  probability distributions to distinguish informative from noisy sentences. This part includes: %We propose two different frameworks in terms of scale of the sentence-level attention in order to (1) indirectly verify the effectiveness of the 2-$D$ word-level attention mechanism, and (2) verify the incremental effectiveness with structured sentence-level attention mechanism: 
%\begin{itemize}
%\item {\bf Strategy 1}: 1-$D$ sentence-level attention which applies to the data where most entity pairs have only one instance;
%\item {\bf Strategy 2}: 2-$D$ sentence-level attention which applies to the data where most entity pairs have more than one instance.
%\end{itemize}

%Regarding {\bf Strategy 1} as shown in Figure~\ref{model_arch}, it includes a 1-$D$ sentence-level attention layer, a 1-$D$ instance selection representation layer and a {\em softmax} output layer. 

%Regarding {\bf Strategy 2}, it includes:
\begin{itemize}
\item {\it a structured sentence-level attention model}: this has a similar structure to the structured word-level attention mechanism, except that it generates a set of summation weight vectors for all input instances in the same bag $\mathcal{G}$. Each vector is a weight distribution over all instances. Accordingly, the 2-$D$ sentence-level matrix is expected to learn a set of different weight distributions focusing on different informative instances. As a result, informative sentences are expected to contribute more with higher weights, and noisy sentences are expected to contribute less with smaller weights, to the relation classification.
\item {\it an averaged sentence-level attention layer}: the 2-$D$  sentence-level attention matrix is averaged and converted to a 1-$D$ vector.

\item {\it a selection representation layer}: the 1-$D$ averaged attention vector is dotted with the output of the flattened representation layer. Accordingly, a 1-$D$ vector, denoted as ``$M_{L2}$'' in Figure~\ref{model_arch}, is obtained which represents an averaged weighted selection representation of multiple sentences. %The difference of this $M_{L2}$ from that in {\bf Strategy 1} is %Each vector in $M_{L2}$ represents a weighted selection of all instances. A  
\item {\it an output layer}: this connects to a {\em softmax} layer and produces a probability distribution corresponding to relation classes.
\end{itemize}   

\subsection{Structured Word-Level Self-Attention and its Penalisation Function}
%The mathematical description of the structured 2-$D$ word-level attention and the structured sentence representation ca be formalised as: 

Given a bag $\mathcal{G}=(S_1, S_2, \ldots, S_J)$ containing $J$ instances, and a sentence $S_j$ in $\mathcal{G}$ consisting of $N$ tokens, $S_j$ can be represented using a sequence of word embeddings, as in~(\ref{eq1}):
\begin{eqnarray}\label{eq1}
 S_j = (e_1, e_2, \ldots, e_N)
\end{eqnarray}

\noindent where $e_i$ is a $d$-dimension vector for the $i$-th word, and $S_j$ is the $j$-th instance in $\mathcal{G}$.

We denote the hidden state of the BiLSTM as in~(\ref{eq2}):
\begin{eqnarray}\label{eq2}
H = ( {\bf h_1, h_2, \ldots, h_N})^T
\end{eqnarray}

\noindent where ${\bf h_t}$ is a concatenation of the forward hidden state $\overrightarrow{h}_{t}$ and the backward hidden state $\overleftarrow{h}_{t}$ at time step $t$. $T$ is the transpose operation. If the size of each unidirectional LSTM is $u$, then $H$ has the size $2u$-by-$N$.

Then, the structured word-level self-attention mechanism is defined as in~(\ref{eq3}):
\begin{eqnarray}\label{eq3}
A_{L1} = softmax(W_{s2}^{L1} tanh(W_{s1}^{L1}H))
\end{eqnarray}
\noindent where $L1$ stands for the first-level attention mechanism, i.e. the word-level; $W_{s1}^{L1}$ is a weight matrix of size $d_a^{L1} \times 2u$, where $d_a^{L1}$ is a hyper-parameter for the number of neurons in the attention network; $W_{s2}^{L1}$ is a weight matrix with the shape $r^{L1}\times d_a^{L1}$, where $r^{L1}$ ($r^{L1}>1$) is the hyper-parameter representing the size of multiple vectors in the 2-$D$ attention matrix. The size of $r^{L1}$ is defined based on how many different aspects of the sentence need to be focused on; $A_{L1}$ is the annotation matrix of size $r^{L1}\times N$. We can see that in $A_{L1}$, there are $r^{L1}$ attention vectors for the $N$-token input sentence.

Finally, we compute the $r^{L1}$ weighted sums by multiplying the annotation matrix $A_{L1}$ and BiLSTM hidden states $H$. The resulting structured sentence representation $M_{L1}$ is~(\ref{eq4}):
\begin{eqnarray}\label{eq4}
M_{L1} = A_{L1}H^T
\end{eqnarray}

\noindent where $M_{L1}$ has the shape $r^{L1}\times 2u$. It can be seen that the traditional 1-$D$ sentence representation is extended to a 2-$D$ representation ($r^{L1} > 1$).

Subsequently, the output of the flattened representation layer for the instance $S_j$ in $\mathcal{G}$ is~(\ref{eq5}):
\begin{eqnarray}\label{eq5}
O^{L1}_j = ReLU(W^{L1}_{o} M_{L1}^{FT} + b_{o}^{L1})
\end{eqnarray}
\noindent where $W^{L1}_{o}$ is the weight matrix that has the shape $v$-by-$r^{L1}*2u$, where $v$ is the amount of neurons in the $ReLU$-based MLP layer; $M_{L1}^{FT}$ is the flattened structured sentence representation which is a concatenated vector of each row in $M_{L1}$ and has the dimension $r^{L1}*2u$; $b_{o}^{L1}$ is the bias vector of size $v$; $O^{L1}_j$ is the aggregated sentence representation of the $j$-th instance in the bag $\mathcal{G}$ with size $v$.

Then, the output of all instances in $\mathcal{G}$ from the flattened representation layer is denoted as in~(\ref{eq6}):
\begin{eqnarray}\label{eq6}
O^{L1} = (O^{L1}_1, O^{L1}_2, \ldots, O^{L1}_J)^T
\end{eqnarray}
\noindent where $O^{L1}$ has the shape of $v\times J$.

As in~\citet{linzhouhan2017}, the penalisation term for the structured word-level attention is as in~(\ref{eq7}):
\begin{eqnarray}\label{eq7}
P_{L1} = ||(A_{L1}A^T_{L1} - I)||_F^2
\end{eqnarray}
\noindent where $||\cdot||_F$ is the Frobenius norm of a matrix. $I$ is an identity matrix. Minimising this penalisation term means that we learn an orthogonal matrix for $A_{L1}$ so that each row in $A_{L1}$ only focuses on a single aspect of semantics.

%\subsection{Strategy 1: 1-$D$ Sentence-Level Attention}
%The 1-$D$ sentence-level attention in {\bf Strategy 1} is calculated as in~(\ref{eq8}):
%\begin{eqnarray}\label{eq8}
%A_{L2} = softmax(W_{a}^{L1} O^{L1})
%\end{eqnarray}

%\noindent where $W_{a}^{L1}$ is a weight vector with the size of $v$, and $A_{L2}$ is the 1-$D$ sentence-level annotation vector with the size of $J$.

%Then, the 1-$D$ sentence selection embedding and the final output for relation prediction in terms of {\bf Strategy 1} can be calculated as in~(\ref{eq9}) and~(\ref{stra1}):
%\begin{align}\label{eq9}
%&M_{L2} = A_{L2}\cdot (O^{L1})^T\\ 
%&p(\hat{y}|S) = softmax(W_o^{L2}tanh(M_{L2})+b_o^{L2}) \label{stra1}
%\end{align}

%\noindent where $M_{L2}$ is the selection representation with the size of $v$, $W_o^{L2}$ is a weight matrix with the dimension of $c\times v$, where $c$ is the number of relation classes. $b_o^{L2}$ is the bias vector with the size of $c$. $p(\hat{y}|S)$ is the probability distribution for $c$ relation classes, and $\hat{y}$ is the predicted relation type.

\subsection{Structured Sentence-Level Self-Attention and Averaged Selection Representation}

%In {\bf Strategy 2}, 
Taking $O^{L1}$ as the input to the structured 2-$D$ sentence-level attention model, the annotation matrix $A_{L2}$ is calculated as in~(\ref{eq11}):
\begin{eqnarray}\label{eq11}
A_{L2} = softmax(W_{s2}^{L2}tanh(W_{s1}^{L2}O^{L1}))
\end{eqnarray}
\noindent where $W_{s1}^{L2}$ is the weight matrix of size $d_a^{L2}\times v$, and $d_a^{L2}$ is the number of neurons in the attention network; $W_{s2}^{L2}$ is the weight matrix of  shape $r^{L2}\times d_a^{L2}$, where $r^{L2}$ ($r^{L2}>1$) is the hyper-parameter representing the size of multiple vectors in the 2-$D$ sentence-level attention matrix. The $r^{L2}$ multiple vectors are expected to focus on different informative instances for the relation classification; $A_{L2}$ is the sentence-level annotation matrix of size $r^{L2}\times J$. We can see that the traditional 1-$D$ sentence-level attention model is expanded to a multi-vector attention ($r^{L2} > 1$).

%Accordingly, we calculate the $r^{L2}$ weighted sums by multiplying the sentence-level annotation matrix $A_{L2}$ and the aggregated sentence representation $O^{L1}$, the resulting structured instance selection representation $M_{L2}$ is:
Then, we average the 2-$D$ $A_{L2}$ to a 1-$D$ vector $\bar{A}_{L2}$ which has the dimension of $J$. %Two reasons that we average the 2-$D$ sentence-level attention rather than concatenate or flatten it are: (1) it 

Accordingly, we calculate the averaged weighted sum by multiplying $\bar{A}_{L2}$ and the aggregated sentence representation $O^{L1}$, with the resulting instance selection representation $M_{L2}$ being~(\ref{eq12}):
\begin{eqnarray}\label{eq12}
M_{L2} = \bar{A}_{L2}\cdot (O^{L1})^T
\end{eqnarray}

\noindent where $M_{L2}$ has the size of $v$. 

%Similarly, the penalisation term $P_{L2}$ can be computed as in Eq.~(\ref{eq5}), and $M_{L2}$ can be flattened to a vector and connects to a $ReLU$-based MLP layer to produce the output vector $O^{L2}$ as:
%\begin{eqnarray}
%O^{L2} = ReLU(W_o^{L2}M_{L2}^{FT} + b_o^{L2})
%\end{eqnarray}
%\noindent where $W_o^{L2}$ has the size of $q\times r^{L2}*v$, where $q$ is the number of neurons in the MLP layer; $M_{L2}^{FT}$ is the flattened representation with the size of $r^{L2}*v$, $b_o^{L2}$ has the dimension of $q$. $O^{L2}$ is a vector with the dimension of $q$.
The probability distribution of the predicted relation type, i.e. the final output for relation prediction, can be calculated as in~(\ref{stra1}):
%\begin{align}\label{eq9}
%&M_{L2} = A_{L2}\cdot (O^{L1})^T\\ 
\begin{eqnarray}
p(\hat{y}|G) = softmax(W_o^{L2}tanh(M_{L2})+b_o^{L2}) \label{stra1}
\end{eqnarray}

%type in terms of {\bf Strategy 2} is the same as in Eq.~(\ref{stra1}). % computed as:
%\begin{eqnarray}\label{eq10}
% p(\hat{y}|S) = softmax(W_o O^{L2} + b_o) 
%\end{eqnarray}
%\noindent where $W_o$ is the weight matrix with the size of $c\times q$, where $c$ is the number of relation classes. $b_o$ is the bias vector with the size of $c$. 
%Note that {\bf Strategy 1} can be regarded as a special case of {\bf Strategy 2} if $r^{L2} = 1$.

\subsection{Loss Function and Optimisation}
%For both Strategies, 
The total loss of the network is the summation of the penalisation term $P_{L1}$, {\em softmax} loss in Eq.~(\ref{stra1}) and the $L2$ regularisation loss.%; For {\bf Strategy 2}, the total loss is the summation of the penalisation terms $P_{L1}$ and $P_{L2}$, $softmax$ loss in Eq.~(\ref{eq10}) and $L2$ regularisation loss.
%Besides two penalisation terms, namely $P_{L1}$ and $P_{L2}$, we also add up the $softmax$ loss in Eq.~(\ref{eq10})  and $L2$ regularisation loss to the final loss to optimize the hyper-parameters.

We use the ADAM optimiser~\cite{adam2014} to minimize the loss function on the mini-batch basis which is randomly selected from the training set. %For learning, we iterate by randomly selecting a mini-batch from the training set until converge.

\section{Experiments}\label{experiments}
\subsection{Datasets}\label{dataset}
%We use the  widely-used data sets to verify the structured self-attention mechanism for the RE task, namely the SemEval 2010 Task 8 (SemEval2010)~\cite{hendrickx2009semeval} and 

We use two distantly supervised datasets, namely the NYT corpus (NYT) and the DBpedia Portuguese dataset (PT),\footnote{There are several reasons to use the Portuguese dataset: (i) the data sets reported in previous work, such as the KBP data, are not publicly available, or (ii) SemEval data sets which are not distantly supervised data. Google has also released a dataset (https://github.com/google-research-datasets/relation-extraction-corpus), but it is smaller and only has 4 relation types. For all these reasons, the Portuguese data is a better option to verify our method.} to verify our method. % with different settings. 
%The SemEval2010 contains 10,717 annotated examples, including 8,000 training instances and 2,717 test instances. There are 9 relationships with two directions and an undirected `Other' class. In terms of the NYT corpus, 
%In this corpus, the data between 2005 and 2006 are labeled as training instances, and the data from 2007 are aligned as testing instances. 

In the NYT dataset, there are 53 relationships including a special relation {\it NA} which indicates a {\it None Relation} between two entities. 
%By analysing the data, we found that the NYT data has two imbalanced issues: (1) 
The training set contains 580,888 sentences, 292,484 entity pairs and 19,429 relational facts (Non-NA). The test set contains 172,448 sentences, 96,678 entity pairs and 1,950 relational facts (Non-NA). %We can see that most entity pairs are labeled as {\it NA}. (2) Most entity pairs have only one instance, i.e. 
There are 19.24\% and 22.57\% entity pairs corresponding to multiple instances in the training set and test set, respectively. %Therefore, two issues mentioned in Section~\ref{ourmodel} -- namely the sentence representation and instance selection -- become extremely important for a model to deal with such imbalanced data in an MIL framework.

The DBpedia Portuguese dataset is smaller, containing just 10 relationships including a special relation {\it Other}. % indicating a {\it None Relation} between two entities. 
%By analysing the data, we found that the NYT data has two imbalanced issues: (1) 
After preprocessing the original dataset, we obtain 96,847 sentences, 85,528 entity pairs and 77,321 relational facts (Non-Other). There are 8.61\% entity pairs corresponding to multiple instances in the whole dataset. As in~\citet{Batista:2013}, we use two different settings for the training and test sets: (1) a manually reviewed subset that contains 602 sentences (PT-MANUAL) as the test set; and (2) 70\%--30\% out of the whole data as the training set and test set, respectively (PT-SPLIT).

%The training set contains 580,888 sentences, 292,484 entity pairs and 19,429 relational facts (Non-NA). The test set contains 172,448 sentences, 96,678 entity pairs and 1,950 relational facts (Non-NA). %We can see that most entity pairs are labeled as {\it NA}. (2) Most entity pairs have only one instance, i.e. 
%There are 19.24\% and 22.57\% entity pairs corresponding to multiple instances in the training set and test set, respectively. 

%As mentioned that predicting the relation category on the NYT corpus is a multi-instance learning problem because some entity pairs have multiple instances and the model should select the informative sentences as much as possible.

 %In this case, we infer that the capability of structured 2-$D$ sentence-level attention might be restricted. %Our experimental results confirm our inference.
%{\color{red}  give the statistics of one-instance and multiple-instance.}

%The SemEval2010 data set is mainly used to verify the structured word-level self-attention mechanism compared with the traditional 1-$D$ attention. The NYT data set is employed to verify the proposed multi-level multi-scale self-attention mechanism on the distant supervised data.
\subsection{Word Embeddings and Relative Position Features}
%As in previous work, we use the 100-dimensional word vectors pre-trained by~\cite{Pennington2014} to initialize the embedding layer for the SemEval2010 task. For the NYT task, 
For the NYT dataset, we use the 200-dimensional word vectors pre-trained using the NYT corpus;\footnote{\url{https://catalog.ldc.upenn.edu/ldc2008t19}} for the PT dataset, we use a pre-trained 300-dimensional word vector model.\footnote{\url{https://s3-us-west-1.amazonaws.com/fasttext-vectors/wiki.pt.vec}} For the two-word entities in the data set, we use \textit{underscore} to connect them as one word. The word embeddings of unknown words are intialised using the normal distribution with the standard deviation 0.05.
Similar to previous work, we also use position embeddings specified by entity pairs. It is defined as the combination of the relative distances from the current word to head or tail entities~\cite{Zeng2014,Zeng2015,Lin2016acl}. %The embedding size of relative position is set to 25.

\subsection{Baselines and Our MLSSA Systems}
Neural RE systems have become the state-of-the-art, such as CNN-based~\cite{Zeng2014,Lin2017acl}, Piecewise CNN-based~\cite{Zeng2015,Lin2016acl,GuoliangJi}, and BiLSTM-based~\cite{Zhoupeng2015} models with or without an attention mechanism. 
In order to carry out a fair comparison, we select CNN+ATT, PCNN+ATT, BiGRU+ATT (bidirectional gated recurrent unit) and BiGRU+2ATT models as baselines on the NYT data, PCNN+ATT and BiGRU+2ATT as baselines on the PT data, where ATT indicates that the model has a sentence-level attention mechanism, and 2ATT indicates that the model has a 1-$D$ word-level and a 1-$D$ sentence-level attention.\footnote{All the baseline systems are obtained from \url{https://github.com/thunlp/NRE} and \url{https://github.com/thunlp/TensorFlow-NRE}.} %In order to show that our proposed MLMS-ATT system has a better capability to learn different aspects of contexts and select more informative sentences, 
%The input to all systems are same, i.e. only containing word embeddings and relation position embeddings without any other external knowledge, such as the multilingual information in~\cite{Lin2017acl}, and entity description~\cite{GuoliangJi}.  %We run these systems by ourselves and achieve comparable results as reported in
% imple- ment them by ourselves which achieve compara- ble results as the authors reported. 

%We use cross-validation to determine the hyper-parameters of our system regarding two different strategies. We name {\bf Strategy 1} as MLMS-ATT-1 and {\bf Strategy 2} as MLMS-ATT-2, respectively.
To show the incremental effectiveness of structured 2-$D$ word-level and 2-$D$ sentence-level self-attention mechanisms, we use two different settings for our MLSSA system: (1) \textbf{MLSSA-1}: this has a 2-$D$ word-level self-attention and a 1-$D$ sentence-level attention, i.e. $A_{L2}$ in Figure~\ref{model_arch} is a 1-$D$ vector. This system is used to verify the context representation learning targeting \textbf{Problem I}; (2) \textbf{MLSSA-2}: both the word-level and sentence-level attentions are structured 2-$D$ matrices. This system verifies the instance selection representation learning targeting \textbf{Problem II}.

\subsection{Experiment Setup and Evaluation Metrics}\label{exp-setup}
Following previous work, we use different evaluation metrics on these two datasets. For the NYT dataset: % 
\begin{itemize}
\item Overall evaluation: all training data is used for the model training, and all test data is used for the evaluation in terms of Precision-Recall (PR) curves;
\item P@N evaluation: we select those entity pairs that have more than one instance to carry out the comparison in terms of the {\it precision} at $n$  (P@N) measure.\footnote{P@N considers only the topmost results returned by the model.} As in~\citet{Lin2016acl}, there are three  settings: (1) One: for each testing entity pair corresponding to multiple instances, we randomly select one sentence to predict the relation; (2) Two: for each testing entity pair with multiple instances, we randomly select two sentences for the relation extraction; and (3) All: for each entity pair having multiple instances, we use all of them to predict the relation. Note that these three selections are only applied to the test set, and we keep all sentences in the training data for model building.
%\item Multiple instance evaluation: our proposed model has merits of learning different aspects of the instance and selecting informative sentences via the 2-$D$ structured attentions, so we select out all entity pairs corresponding to more than 10 instances from the original training data as a new training set, and from the original test set as a new test set, respectively, and compare with baselines in terms of PR curves.
\end{itemize}

For the PT dataset, we use Macro F1 to evaluate system performance.\footnote{Regarding the metric, we keep the evaluation consistent with the work in~\citet{Batista:2013} where they used F1 to measure their RE systems on the Portuguese dataset, in order to maintain a fair comparison with their work using the same metric.}

\subsection{Hyper-parameter Settings}
We use cross-validation to determine the hyper-parameters of our system regarding two different settings and datasets. The in-common and different parameters for our two systems and two datasets are shown in Table~\ref{tab-paras}.
\begin{table}[h]
\small
\centering
%\begin{tabular}{p{2.5cm}p{3cm}p{1.2cm}}
\begin{tabular}{l|r|r}
\hline
\textbf{Parameters for MLSSA-1/2} & \textbf{NYT} &\textbf{PT} \\
\hline
Word embedding dimension $d$ & 200 &300 \\ \hline
Position embedding dimension & 50 & 50 \\ \hline
Batch size $B$ & 64 & 50\\ \hline
Time steps $T$ & 70 & 70\\ \hline
Learning rate $\lambda$ & 0.001 & 0.001 \\ \hline
%Dropout probability $p$ & 0.2 \\ \hline
Hidden size in BiLSTM $u$ & 300 & 300 \\ \hline
$d_a^{L1}$ at word-level attention & 300 & 300 \\ \hline
$r^{L1}$ at word-level attention & 9 & 5\\ \hline
MLP size $v$ & 1000 & 1000\\ \hline

Coefficient of the penalisation term & 1.0 & 1.0 \\ \hline \hline
{\bf Parameters for MLSSA-2 only} & {\bf NYT} & \textbf{PT} \\ \hline
$d_a^{L2}$ at sentence-level attention & 300 & 300 \\ \hline
$r^{L2}$ at sentence-level attention & 9 & 3\\ \hline
%MLP size $q$ of the output layer & 1000 \\ \hline
%79.32\\ \hline
%Simple & 67.32 \\
%Simple + POS &  68.87\\
%Simple + POS + Dep & 71.35\\
%Simple + POS + Dep + NE & 71.71\\
%\hline
\end{tabular}
\caption{Hyper-parameter settings}
\label{tab-paras}
\end{table}

%Note that in our systems, Dropout is not used.

\begin{table*}[tbp]
\small
\centering
%\begin{tabular}{p{2.5cm}p{3cm}p{1.2cm}}
\begin{tabular}{l|c|c|c|c||c|c|c|c||c|c|c|c}
\hline
\textbf{Test Settings} &\multicolumn{4}{c||}{\textbf{One}}& \multicolumn{4}{c||}{\textbf{Two}} &\multicolumn{4}{c}{\textbf{All}} \\
\hline
P@N(\%) & 100 & 200 & 300 & Mean & 100 & 200 & 300 & Mean & 100 & 200 & 300 & Mean \\ \hline
CNN+ATT & 72.0 & 67.0 & 59.5 & 66.2 & 75.5 & 69.0 & 63.3 & 69.3 & 74.3 & 71.5 & 64.5 & 70.1 \\ \hline
PCNN+ATT & 73.3 & 69.2 & 60.8 & 67.8 & 77.2 & 71.6 & 66.1 & 71.6 & 76.2 & 73.1 & 67.4 & 72.2 \\ \hline
BiGRU+ATT & 75.0 & 69.5& 64.7 & 69.7 & 80.0 & 72.5& 69.3 &73.9 & 82.0 & 76.5& 71.3& 76.6\\ \hline
%BiGRU+ATT & 75.0 & 69.5& 63.0 & 69.2 & 70.0 & 70.0& 65.0 & 68.3& 71.0 & 73.0& 69.0& 71.0\\ \hline
%BGRU+2ATT & 75.0 & 70.5 & 63.3 & 69.6 & 75.0 & 71.5 & 66.7 & 71.1 & 80.0 & 75.0 & 71.0 & 75.3 \\ \hline
BiGRU+2ATT & 81.0 & 74.0 & 67.3 & 74.1 & 81.0 & 75.5 & 70.7 & 75.7 & 81.0 & 76.0 & 72.7 & 76.6 \\ \hline \hline
%ML-SSA & {\bf 79.0} & {\bf72.5} & {\bf63.0} & {\bf71.5} & {\bf82.0} & {\bf77.5} & {\bf68.7} & {\bf76.1} & {\bf84.0} & {\bf76.5} & {\bf71.7} & {\bf77.4}\\ \hline

%MLMS-ATT-1 & 84.0 & 75.5 & {70.3} & { 76.6 } & {82.0} & {76.0} & {71.3} & { 76.4 } & {86.0} & {79.5} & {76.3} & { 80.6}\\ \hline
%MLMS-ATT-2 & {\bf 85.0} & {\bf 77.5} & {\bf 70.3} &{\bf 77.6} & {\bf 85.0} & {\bf 80.5} &{\bf 71.3} & {\bf 78.9 } & {\bf 86.0} & {\bf 81.0} & {\bf 77.0} & {\bf 81.3}\\ \hline
%model 10200
MLSSA-1 & {\bf 88.0} & {\bf 77.0} & {\bf 70.0} & {\bf 78.3 } & { 88.0} & {\bf 79.0} & {\bf 73.3} & {\bf 80.1 } & { 87.0} & {\bf 81.5} & {76.0} & { 81.5 }\\ \hline
% model: 13700
MLSSA-2 & 87.0 & 76.0 & {\bf 70.0} & 77.7 &  {\bf 89.0} & { 78.5} & 72.3 & 79.9 & {\bf 90.0} & {\bf 81.5} & {\bf 77.0} & {\bf 82.8}\\ \hline
%MLMS-ATT-2 & 85.0 & {\bf 77.5} & {\bf 70.3} & 77.6 &  85.0 & {\bf 80.5} & 71.3 & 78.9 & 86.0 & 81.0 & {\bf 77.0} & 81.3\\ \hline
%MLMS-ATT-2 & {\bf 87.0} & {\bf 77.0} & {\bf 68.7} &{\bf } & {\bf 87.0} & {\bf 77.0} &{\bf 71.6} & {\bf  } & {\bf 86.0} & {\bf 81.0} & {\bf 77.0} & {\bf 81.3}\\ \hline

\end{tabular}
\caption{Precision values for the top-100, top-200, and top-300 relation instances that are randomly selected in terms of one, two and all sentences.}
\label{p_at_n}
\end{table*}
\subsection{PR Curves on NYT Dataset}
\begin{figure}[htbp]
\centering
\includegraphics[scale=0.5]{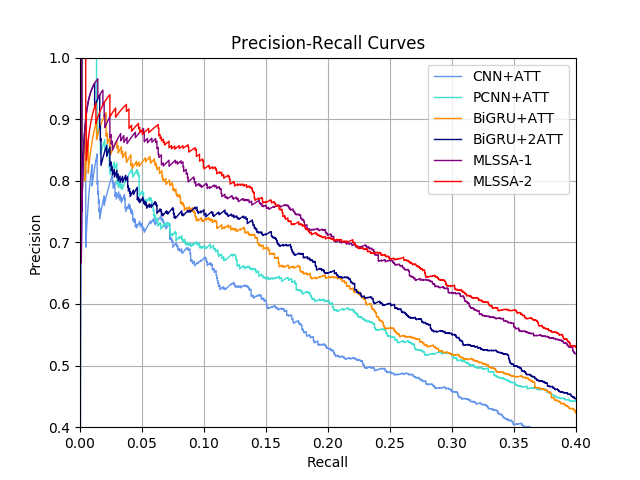}
\caption{Comparison results of a variety of methods in terms of precision/recall curves.}
\label{pr_curve}
\end{figure}

The comparison results for the NYT test set are shown in Figure~\ref{pr_curve}.
We have the following observations: 
%(1) MLMS-ATT-2 performs worst than other systems. The reason is that most entity pairs have only one instance (c.f. Section~\ref{dataset}), so the 2-$D$ sentence-level attention has no choice to select other instances, resulting that the penalisation term $P_{L2}$ cannot be optimised to a diagonal matrix, but introduces more noise. We observed that during the training, $P_{L2}$ is scarcely convergent. %Therefore, we use MLMS-ATT-1 as our main system to compare with baselines. %in terms of {\it Overall Evaluation} and {\it P@N Evaluation} experiments.
%In future work, we will build a real multi-instance data set to verify the effectiveness of {\bf Strategy 2}.
(1) BiGRU+ATT outperforms CNN+ATT and PCNN+ATT in terms of the PR curve, showing that it can learn a better semantic representation from the sequential input;
(2) BiGRU+2ATT has better overall performance compared to BiGRU+ATT, showing that word-level attention is beneficial to sentence-level attention compared to single-attention models, i.e. the sentence-level attention model can select more informative sentences based on a more reasonable sentence embedding learned by the word-level attention model; 
(3) %Although starting from {\it Recall=0.2}, 
MLSSA-1 outperforms all baseline systems in terms of the PR curve, which demonstrates that the structured 2-$D$ word-level attention model can learn a better sentence representation by focusing on different aspects of the sentence, so that the sentence-level attention has a better chance of selecting the most informative sentences; and (4) the PR curve of MLSSA-2  is  higher than that of MLSSA-1, demonstrating that the 2-$D$ sentence-level attention model can better select the most informative sentences compared to the 1-$D$ sentence-level attention model targeting those entity pairs with multiple instances. 
%However, we note that due to the fact that most entity pairs in the NYT data have only one instance (cf. Section~\ref{dataset}), the performance of the 2-$D$ sentence-level attention model might be affected by those entity pairs to some extent.%, i.e. introducing some noise to the optimisation of weight matrices in Eq.~(\ref{eq11}). %Therefore, in our {\bf Strategy 2}, averaging the 2-$D$ $A_{L2}$ to a vector can reduce the noise, and make it more smooth, compared with the method of concatenating  each row in $A_{L2}$. 
%In future work, we will construct a pure multi-instance data set and further verify the effectiveness of 2-$D$ sentence-level attention.%We infer that it might be affected by the unbalance problem of multiple instances, i.e. only a small proportion of entity pairs have multiple instances, so the the optimisation of structured 2-$D$ sentence-level attention wight matrices is  %more {\bf NA} entity pairs with multiple instances are predicted as {\it Non-NA} classes.
%(4) The results show that 

\subsection{P@N Evaluation on NYT Dataset}
The results on the NYT dataset regarding P@100, P@200, P@300 and the mean of three settings for each model are shown in Table~\ref{p_at_n}.
From the table, we have similar observations to the PR Curves: %(1) Our MLMS-ATT-2 performs worst in most cases because of the one-instance learning problem. 
(1) BiGRU+2ATT outperforms CNN+ATT, PCNN+ATT and BiGRU+ATT in most cases in terms of all P@N scores; and (2)  MLSSA-1 and MLSSA-2 significantly outperform all baselines for all measures. We observe that MLSSA-1 performs better than MLSSA-2 on tasks {\bf One} and {\bf Two}, but worse on {\bf All}. %We analyse that it is also caused by the issue that most entity pairs have only one instance. 
We infer that in our 2-$D$ sentence-level attention model, we set $r^{L2}$ to 9, but there are only {\it one} and {\it two} instances for selection in tasks {\bf One} and {\bf Two}, so the 2-$D$ matrix cannot demonstrate its full potential. However, in {\bf All}, many entity pairs contain multiple or more than 9 instances, so it can learn a better 2-$D$ matrix to focus on different instances.% the more instances there are for one entity pair, the more effective it is to select informative sentences. %{\bf All} contains all multiple instances, while {\bf One} and {\bf Two}.%the prediction capability is relatively weak on {\bf One} and {\bf Two} instances compared to {\bf All} instances. % performs best in most P@N measures compared to MLMS-ATT-2, showing that the combination of structured 2-$D$ word-level attention and 1-$D$ sentence-level attention is more effective and robust on the NYT data for the multi-instance learning  and selecting informative sentences.

\subsection{Results on PT Dataset}
Based on results from the NYT dataset, we choose PCNN+ATT and BiGRU+2ATT as representative baselines to compare against our MLSSA-1/2 systems on the PT test sets. The results in terms of Macro F1 are shown in Table~\ref{pt-result}.

It can be seen that on both test sets, our MLSSA-2 model achieved the best performance which shows that the structured 2-$D$ word-level and sentence-level self-attention models can be well applied to datasets of a smaller scale and with a smaller ratio of multiple instances. 

\begin{table}[h]
\small
\centering
%\begin{tabular}{p{2.5cm}p{3cm}p{1.2cm}}
\begin{tabular}{l|c|c}
\hline
SYS & PT-MANUAL (\%) & PT-SPLIT (\%) \\ \hline
%CNN+ATT & & \\ \hline
PCNN+ATT & 62.3& 74.1\\ \hline
%BiGRU+ATT & & \\ \hline
BiGRU+2ATT & 63.5& 75.3\\ \hline\hline
MLSSA-1 & 66.0	 & 77.2\\ \hline
MLSSA-2 & \textbf{69.6}& \textbf{78.1}\\ \hline
\end{tabular}
\caption{Results on the PT test sets}
\label{pt-result}
\end{table}

\begin{figure*}[h]
\centering
\includegraphics[scale=0.4]{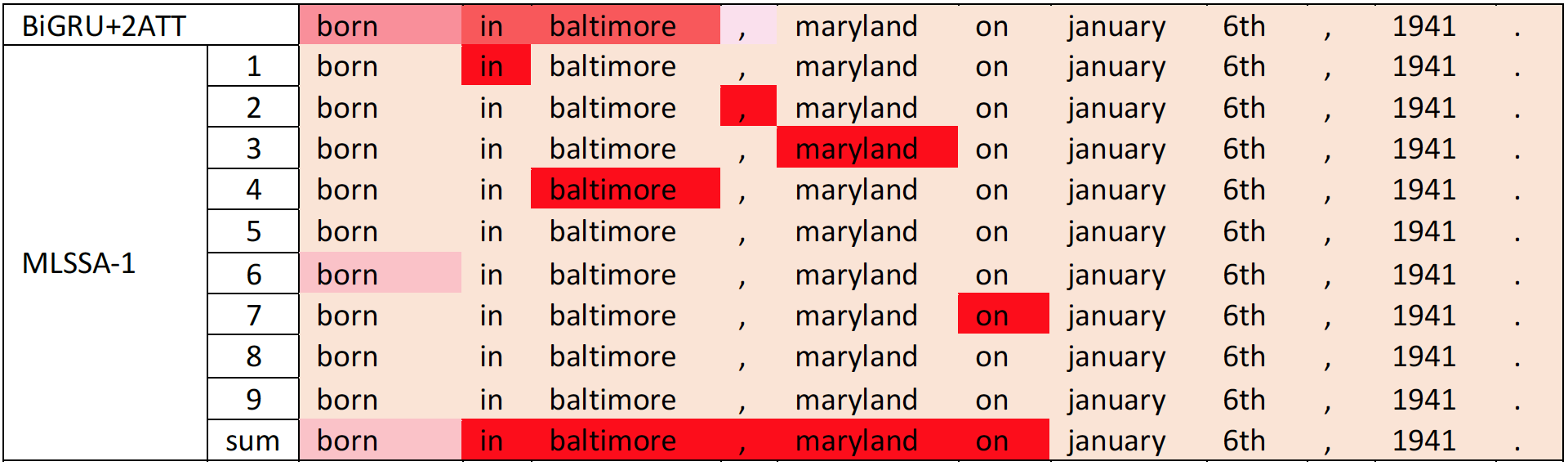}
\caption{Comparison of word-level attentions.}
\label{heatmap_word_level}
\end{figure*}

\subsection{Examples and Analysis}\label{analysis}
In order to show the effectiveness of structured self-attention mechanisms, we show some examples by visualising the attentions on different aspects of a sentence, and on different sentences comparing with BiLSTM+2ATT model.

Figure~\ref{heatmap_word_level} shows the comparison of word-level attention mechanism between BiGRU+2ATT and MLSSA-1 reflecting their capability of context representation learning ({\em Problem I}). MLSSA-2 has a similar probability distribution to MLSSA-1 in terms of this example.

The {\it pink} fonts indicate lower probability and {\it red} indicates higher probability. We observe that: (1) BiGRU+2ATT mainly focuses on one word {\it baltimore}. %, and then the words {\it in} and {\it born}. 
We can see that it has little attention on the entity word {\it maryland}. In this example, the comma implies a semantic relationship {\it location/location/contains} for the entity pair ({\it Maryland, Baltimore}). However, BiGRU+2ATT allocates  quite a small probability to it;   %it is a uniform-like distribution for words in BiGRU+2ATT.  As is known that GRU has the property of memorising the context of the sequence, and the final hidden state often contains a complete context of the input sequence. Accordingly, in the results of the BiGRU+ATT, we discovered that most word-level attentions are on the {\it BLANK} tokens of last time steps. 
 and (2) we can see that our model focuses on different words via different attention vectors (9 in total). Words with a {\it red} background have a high probability of $0.98$ or so. For rows 5, 6, 8 and 9, the focus is on the {\it BLANK} tokens. In both systems, the maximum time step is set to 70, which indicates that shorter sentences are padded with {\it BLANK} tokens and longer sentences are cut off. The last row shows the summation of {\bf 9} annotation vectors, and it constructs a dependency-like context of the relation for the entity pair. % ({\it Maryland, Baltimore}) as {\it location/location/contains}. 
Attentions on different words are attributed to the penalisation  $P_{L1}$ which is optimised to learn orthogonal eigenvectors.
\begin{figure*}[h]
\centering
\includegraphics[scale=0.4]{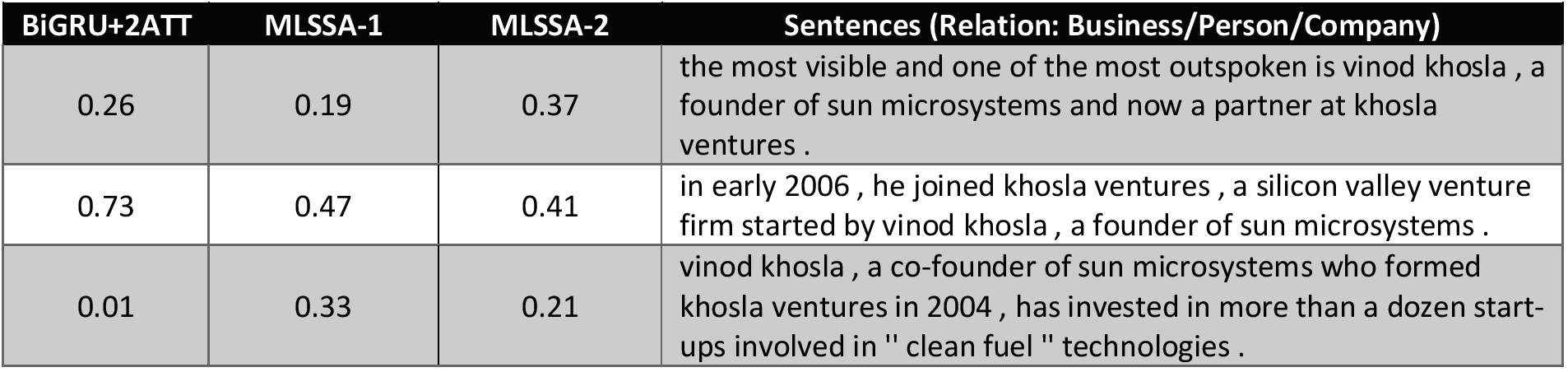}
\caption{Comparison of sentence-level attentions.}
\label{example_sen_level}
\end{figure*}

Figure~\ref{example_sen_level} shows the comparison of sentence-level attentions between BiGRU+2ATT, MLSSA-1 and MLSSA-2. The first, second and third columns are probability distributions over multiple instances. The entity pair is ({\it vinod khosla, sun microsystems}), and their relation is {\it Business/Person/Company}. From this figure, we observe that: (1) BiGRU+2ATT allocates high probabilities to {\bf Sentences 1} and {\bf 2} by learning the context of ``{\it a founder of}'', but does not recognise that ``{\it co-founder}'' is semantically the same as ``{\it founder}''; and (2) our two models almost evenly focus on all sentences because they express the same semantic concept of ``{\it a person is a founder of a company}'' in terms of the given entity pair. Therefore, the structured self-attention mechanism is helpful to learn a better representation and select  informative sentences. 

\section{Conclusion and Future Work}\label{conclusion}
This paper has proposed a multi-level structured self-attention mechanism for distantly supervised RE. In this framework, the traditional 1-$D$ word-level and sentence-level attentions are extended to 2-$D$ structured matrices which can learn different aspects of a sentence, and different informative instances. Experimental results on two distant supervision data sets show that (1) the structured 2-$D$ word-level attention can learn a better sentence representation; (2) the structured 2-$D$ sentence-level attention and averaged selection can perform better selection from multiple instances for relation classification; (3) the  proposed framework significantly outperforms state-of-the-art baseline systems for a range of different measures, which verifies its effectiveness on two representation learning issues. A subsequent manual investigation via examples also show its effectiveness on two representation learning issues.%We report experimental results on NYT data set as: (1) the proposed {\bf Strategy 1} significantly outperforms all baselines in terms of PR curve and P@N evaluations; (2) {\bf Strategy 2} performs worse than expected due to the fact that most entity pairs in the NYT corpus have only one instance, so that the structured 2-$D$ sentence-level attention cannot be optimised.

In future work, we will build a domain-specific distant supervision dataset with a higher ratio of multiple instances  and compare our system with others. 
Furthermore, %based on the analysis in Section~\ref{analysis}, 
we will consider not using RNNs or CNNs, but a deeper neural networks with only attentions for distantly supervised RE, similar to the work in~\citet{Vaswani2017}.

\section*{Acknowledgments}
We thank Yuyun Huang and Utsab Barman from University College Dublin, and Jer Hayes, Edward Burgin and other colleagues from Accenture Labs Dublin for helpful comments, discussion and facilities. We would like to thank the reviewers for their valuable and constructive
comments and suggestions.  This research is supported by the ADAPT Centre for Digital Content Technology, funded under the SFI Research Centres Programme (Grant 13/RC/2106), and by SFI Industry Fellowship Programme 2016 (Grant 16/IFB/4490), and is supported by Accenture Labs Dublin. 

\bibliography{emnlp2018}
\bibliographystyle{acl_natbib_nourl}

\end{document}